\pdfoutput=1
\documentclass[11pt]{article}
\usepackage{preamble}

\usepackage{tikz}
\usepackage{forest}
\usepackage{adjustbox}
\usepackage{caption}
\usepackage{tikz}
\usetikzlibrary{positioning,arrows.meta}
\usepackage{xcolor}
\usepackage{multirow}
\usepackage{enumitem}

\definecolor{roundOne}{HTML}{CAE9D7}   
\definecolor{roundTwo}{HTML}{FFDEB4}   
\definecolor{roundThree}{HTML}{F8C6C6} 

\newcommand{\swallow}[1]{}

\title{Evaluating Language Translation Models by Playing Telephone}

\author{
Syeda Jannatus Saba \quad Steven Skiena \\
Stony Brook University \\
\texttt{\{sysaba, skiena\}@cs.stonybrook.edu} \\
}

\begin{document}
\maketitle
\begin{abstract}
Our ability to efficiently and accurately evaluate the quality of machine translation systems has been outrun by the effectiveness of current language models---which limits the potential for further improving these models on more challenging tasks like long-form and literary translation.
We propose an unsupervised method to generate training data for translation evaluation over different document lengths and application domains by repeated rounds of translation between source and target languages.
We evaluate evaluation systems trained on texts mechanically generated using both model rotation and language translation approaches, demonstrating improved performance over a popular translation evaluation system (\texttt{xCOMET}) on two different tasks: (i) scoring the quality of a given translation against a human reference and (ii) selecting which of two translations is generationally closer to an original source document.
\end{abstract}

\section{Introduction}
\label{sec:introduction}

Machine translation (MT) has become a fundamental pillar of multilingual communication, enabling rapid and scalable information transfer across languages.
However, our ability to efficiently and accurately evaluate the quality of machine translation systems has been outrun by the effectiveness of current language models---which limits the potential for further improving these models on more challenging tasks like long-form and literary translation.
Accurate translation evaluation is needed to inform model development, assess training strategies, and help with real-world deployment decisions in commercial applications. 

Properly evaluating translation systems is a complex task, balancing issues of correctness and readability.
Multiple valid translations exist for every given text, and proper assessment requires capturing nuances in semantics, style, and fluency. Standard metrics like BLEU~\cite{papineni2002bleu} rely heavily on n-gram overlap with a single human reference translation, often penalizing semantically correct yet stylistically or structurally divergent translations.
Recent neural metrics such as \texttt{COMET}~\cite{rei2020comet} and \texttt{xCOMET}~\cite{guerreiro2024xcomet} use machine learning to train high-quality evaluators from human-annotated reference translations.
However, human annotation is costly, and requires specialized expertise to prepare and score gold standard translations, particularly between pairs of low resource languages.
These evaluation techniques will not scale as the research frontier in translation advances from single-sentence translation to the greater challenges of technical and literary translation of poetry, full-length narratives, and textbooks.

\begin{figure*}[t]
  \centering
  \includegraphics[width=\linewidth]{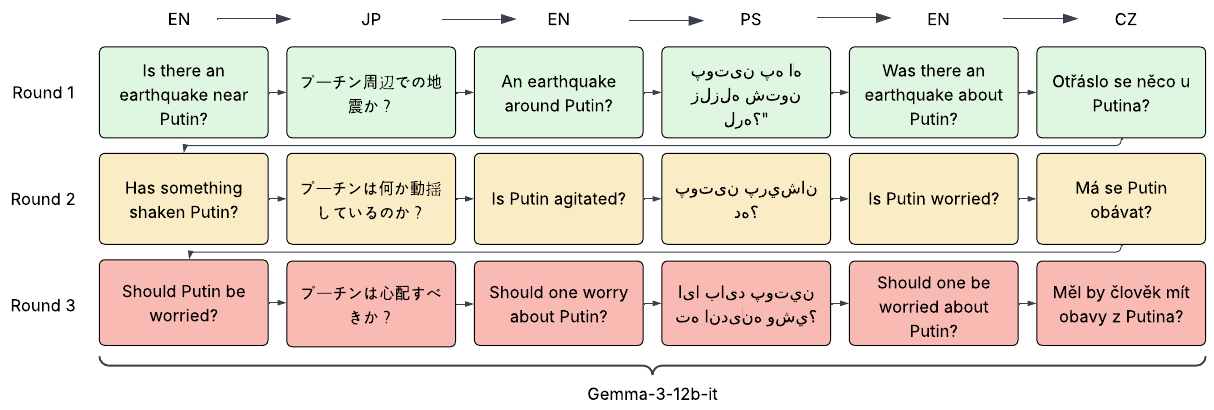} 
  \caption{Language rotation setup for iterative translation. Starting from the Czech sentence ``Zemětřesení kolem Putina?'', we rotate through Japanese and Pashto. Meaning gradually shifts from a factual query to subjective concern.
\vspace*{-1.5em}
}
  \label{fig:language-rotation-example}
\end{figure*}

In this work, we propose a novel unsupervised method to generate training data for translation evaluation over different document lengths and application domains---without requiring any human annotation.
Our approach is inspired by the popular children's game ``Telephone''.
Here the children playing order themselves in a line, and the first child makes up a message they whisper to the second child.
This process repeats until the end of the line, and with each stage in transmission some fidelity is usually lost.
Merriment ensues when the last child reveals what they think the initial message was.

We leverage this idea to generate training data for language translation.
Starting from an initial text, we use existing language translation models through repeated rounds of translation between source and target languages.
Just as with the children's game, each successive translation can be assumed to be more corrupted with respect to the original source than its predecessor, providing a natural measure of translation quality without human annotation.
Careful analysis of such translation sequences allows us to produce pseudo-labels that capture both variations in intrinsic sentence difficulty and the dynamics of semantic drift over multiple translation cycles.

Our primary contributions include:
\begin{itemize}
    \item {\em Translation evaluation without human annotation}:  We introduce our ``Game of Telephone'' approach, providing an unsupervised pipeline to generate unlimited training data from iterative translations scored by existing automatic metrics. 
    This method 
    substantially reduces the barrier to developing robust MT evaluation models, especially for low-resource language scenarios where human annotations are scarce or unavailable.

    In the reference evaluation task (given a source text in language $L_1$ and its reference translation in another language $L_2$, score a second translation into $L_2$), our telephone-trained model outperformed (0.604) the state-of-the-art \texttt{xCOMET}~\cite{guerreiro2024xcomet} in Pearson correlation (0.514) against the human reference standard.
    

    \item {\em Language rotation vs.\ model rotation}: We systematically explore two variants of our approach: language-rotated translation cycles (varying the pivot languages at each iteration) and model-rotated translation cycles (varying MT models at each iteration). We empirically demonstrate both strategies can capture nuanced translation degradation: both language- and model-rotation trained models obtained higher AUC scores than \texttt{xCOMET} at predicting the earlier generation texts for five different LLMs.

    \item {\em New translation evaluation benchmark}: We have released evaluation metrics (both code and data) to assess translator systems for sensitivity to semantic drift as well as sentence-level accuracy/robustness\footnote{https://github.com/saba-phoenix/language-translation-evaluation-by-playing-telephone}. This serves as a resource for advancing language translation quality assessment methodologies.

\end{itemize}

Our paper is organized as follows.
We survey previous work in translation evaluation in Section \ref{sec:previous_work}.
Details of our approach to quality scoring as a function of translation round, document difficulty, and relative model fidelity are discussed in Section \ref{sec:scoring_telephone}.
We explain how we use this data to train evaluation models in Section \ref{sec:training}, with our experimental results in Section \ref{sec:experimental_results}.
We conclude with ideas for future research in Section \ref{sec:discussion}.

\section{Previous Work}
\label{sec:previous_work}

\paragraph{Machine Translation.}
MT has advanced with large transformers—from the original Transformer~\cite{vaswani2017attention} to multilingual models like mBART~\cite{liu2020multilingual}, M2M-100~\cite{fan2021beyond}, and NLLB-200~\cite{nllb2022}. Instruction-tuned LLMs such as GPT-4~\cite{openai2023gpt4}, Mixtral~\cite{mistral2024mixtral}, and Gemini~\cite{google2023gemini} now handle translation as a byproduct of language understanding. Still, low-resource performance lags~\cite{ruder2021xtreme, freitag2022high}, and issues like idioms and domain shift demand better evaluation and adaptation.

\begin{figure*}[t]
  \centering
  \includegraphics[width=\linewidth]{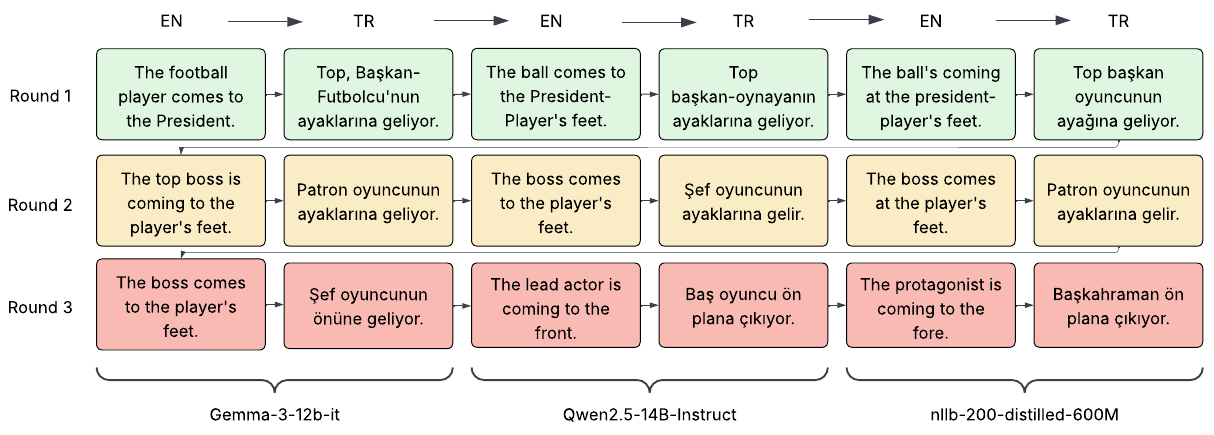} 
  \caption{Model rotation for iterative translation.
We perform three rounds of Turkish-English round-trip translations on  “Futbolcu Başkan’ın ayağına gelir”. Later iterations reveal compounding semantic drift.\vspace*{-1.2em}}
  \label{fig:model-rotation-example}
\end{figure*}

\paragraph{Translation Evaluation System.} While MT models have advanced rapidly, evaluation metrics have lagged behind. Traditional metrics such as BLEU~\cite{papineni2002bleu}, METEOR~\cite{banerjee2005meteor}, and TER~\cite{snover2006study} rely on surface-level n-gram overlap, frequently failing to capture semantic adequacy or fluency. This gap has driven the development of learned metrics more aligned with human judgment, such as \texttt{COMET} and \texttt{BLEURT}~\cite{sellam2020bleurt}. Benchmarks like WMT~\cite{freitag2021results}, MLQE-PE~\cite{fomicheva2020unsupervised}, and FLORES~\cite{nllbteam2022} have become central to evaluating translation quality.

\vspace*{-0.5em}
\paragraph{COMET and xCOMET.} Learned metrics like \texttt{COMET} and its extensions \texttt{COMETKIWI}~\cite{rei2022cometkiwi}, \texttt{xCOMET} have improved MT evaluation by aligning closely with human judgments, incorporating uncertainty estimates~\cite{glushkova2021uncertainty} and span-level error detection. Yet recent findings~\citep{agrawal2023can} reveal a saturation effect at the upper end of the quality spectrum, where these metrics struggle to distinguish between strong translations. This is particularly limiting when comparing top-tier systems or reranking hypotheses. 

\paragraph{Synthetic Data Generation.} \citet{tuan2021quality} create synthetic QE data by applying MT and MLM rewriting over mined parallel corpora. The main idea is to inject noise into clean translations, and then recover word-level tags through edit distance with a pseudo-reference. \citet{etchegoyhen2023learning} collect translations from MT models at different training checkpoints, using earlier checkpoints to simulate weaker systems and pairing them with final model outputs as references.

\paragraph{Translation Chains and Semantic Drift. }Beyond static evaluations, real-world applications often involve translation chains,
such as multilingual pivots or model rotations, where quality can drift gradually. Prior work on cyclic translation obfuscation~\citep{potthast2013overview} formalizes this idea, using multi-hop translations across diverse languages and engines to paraphrase content while preserving meaning. Since different intermediate languages reorder, compress, or expand information differently~\citep{beinbornchoenni2020semantic}, such paths naturally introduce subtle distortions. Existing metrics overlook these dynamics.
To address these gaps, we construct a degradation-aware scoring framework built on iterative translations. We now describe how we transform raw metric outputs into more meaningful learning signals.

\section{Scoring Telephone Translations}
\label{sec:scoring_telephone}

In our Telephone setup, a sentence is translated through multiple rounds of MT, each introducing potential semantic drift through model- or language-rotation. This setting exposes two hidden factors that influence translation quality:

\begin{itemize}[noitemsep, topsep=1pt]
    \item \textbf{Global sentence difficulty:} Some sentences degrade quickly regardless of the iteration or system used.
    \item \textbf{Local iteration pressure:} Some iterations are inherently harder due to the model or pivot language used.
\end{itemize}

Standard metrics like \texttt{xCOMET} overlook these, treating outputs as independent and context-free. We transform raw metric scores into a more informative, context-aware scoring signal by accounting for sentence fragility and iteration difficulty, yielding a refined score that tracks translation quality without human supervision.

\subsection{Score Normalization Procedure}
Formally, given a source sentence $s_i$ from a dataset of $N$ sentences, and its translation $T_i^{(j)}$ at iteration $j$, we compute raw scores $q_i^{(j)}$ using an automatic metric (\texttt{xCOMET}), and aim to transform them into context-aware scores $r_i^{(j)}$.

\paragraph{Step 1: Estimate Sentence Fragility.}To capture sentence-specific intrinsic translation difficulty, we define a sentence-level hardness measure by averaging each sentence's quality scores over all $K$ iterations and standardize it across the dataset:
\vspace*{-0.5em}

\begin{equation}
\quad z_i = \frac{\mu_i - \bar{\mu}}{\bar{\sigma}}
\vspace*{-0.3em}
\end{equation}

where $\bar{\mu}$ and $\bar{\sigma}$ are the mean and standard deviation of $\mu_i$ across the dataset. This $z_i$ serves as a global indicator of how inherently difficult sentence $s_i$ is to translate.

 \paragraph{Step 2: Estimate Iteration Pressure.} 
We characterize each iteration's overall translation quality profile by computing iteration-specific statistics, namely the mean $\mu^{(j)}$ and standard deviation $\sigma^{(j)}$ across the raw scores of all sentences at iteration $j$.

\begin{equation}
\mu^{(j)} = \frac{1}{N}\sum_{i=1}^{N} q_i^{(j)}
\end{equation}
\begin{equation}
\sigma^{(j)} = \sqrt{\frac{1}{N}\sum_{i=1}^{N}(q_i^{(j)} - \mu^{(j)})^2}
\end{equation}

where $N$ is the number of sentences.

\paragraph{Step 3: Compute Refined Score.} Finally, we produce the refined, context-aware translation score $r_i^{(j)}$ by re-projecting each sentence's global difficulty $z_i$ into the local score distribution at iteration $j$:
\vspace*{-0.5em}
\begin{equation}
r_i^{(j)} = \mu^{(j)} + z_i \cdot \sigma^{(j)}
\vspace*{-0.5em}
\end{equation}


This cross-distribution projection yields a refined score $r_{i}^{(j)}$ that is designed to capture how difficult sentence $s_i$ is to translate under the pressure of iteration $j$. 
The transformation helps disentangle raw metric scores from local and global confounders, producing a degradation-aware signal suitable for evaluator training or analysis.

\section{Training Evaluation Models}
\label{sec:training}

\subsection{Training Data Construction}
We use 114,864 source-reference pairs from WMT DA (2018--2022)~\cite{kocmi2022findings}, across 36 language pairs, and split them 80/20 into training and validation. 
We build iterative translation chains where each sentence undergoes 18 translation rounds. This number arises from our design: 3 rotation setups (model or language triplets) $\times$ 2 directions of translation per iteration (forward and back) $\times$ 3 iterations per setup = 18. The choice ensures (i) $\sim$1M synthetic examples, comparable to the ~1M DA-labeled examples used in xCOMET pretraining, and (ii) at least three complete forward iterations per setup, which are required for our paired-generation experiments in Subsection~\ref{subsec:paired_generation}.

\begin{figure}[h]
    \centering
    \includegraphics[width=0.99\linewidth]{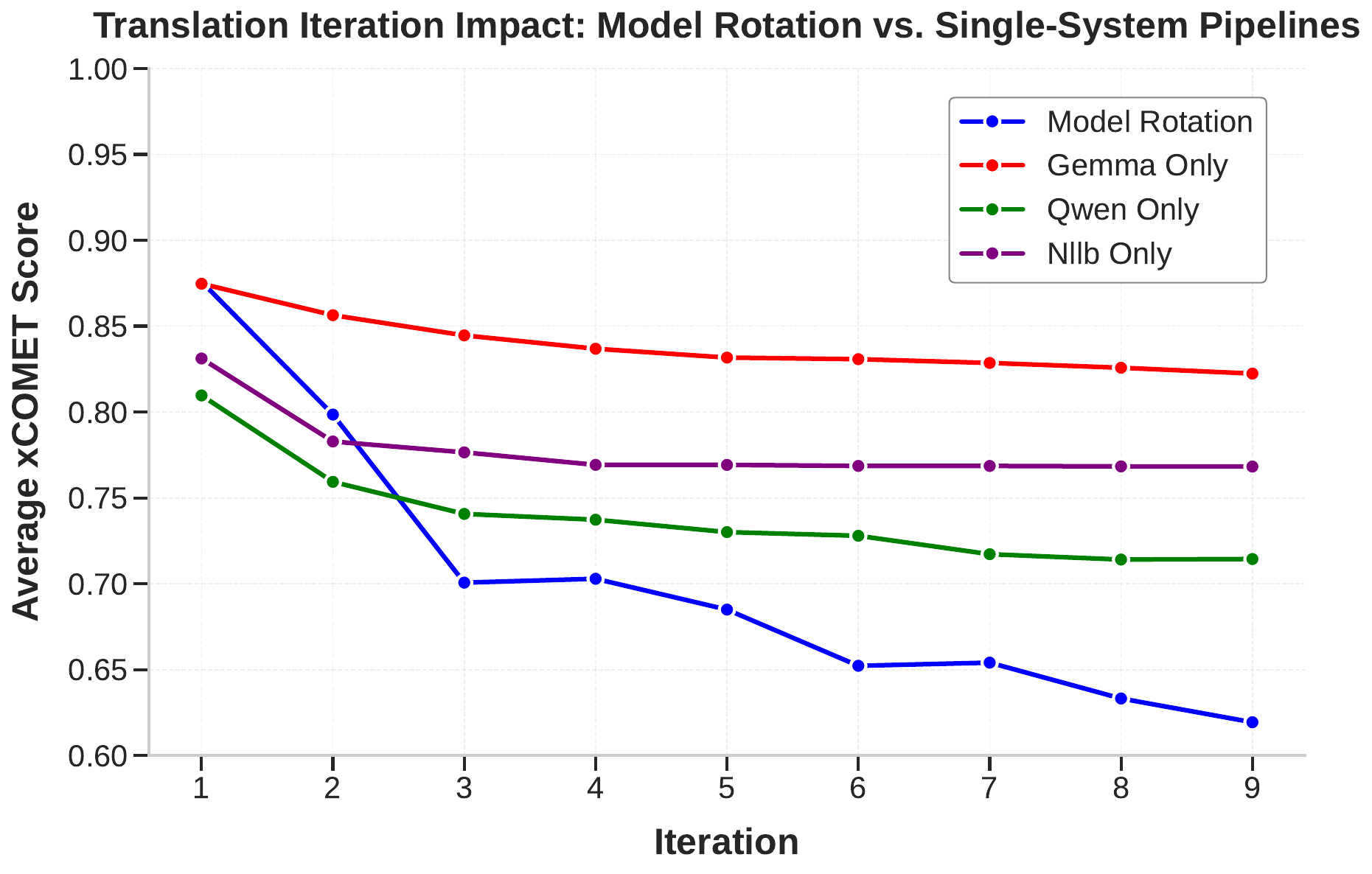} 
    \caption{Average \texttt{xCOMET} scores over forward 9 iterations for 300 sentences. Model rotation degrades fastest; single-model pipelines plateau after early decline.
    \vspace*{-1em}
}
    \label{fig:degradation}
\end{figure}

We consider two complementary rotation strategies for constructing these chains:

\paragraph{Model Rotation.}
In this setup (Figure~\ref{fig:model-rotation-example}), we fix the language pair 
and rotate the MT model at each iteration. Each sentence is translated and back-translated using one model, then passed to the next in a cyclic order. This avoids convergence artifacts like idempotency and introduces diverse distortions that accelerate semantic drift. As shown in Figure~\ref{fig:degradation}, model rotation leads to faster degradation compared to static systems.

Our training rotation involves three distinct MT systems: \texttt{gemma-3-12b-it}~\cite{anil2024gemma}, \texttt{Qwen2.5-14B-Instruct}~\cite{cui2024qwen2}, and \texttt{nllb-200-distilled-600M}~\cite{nllb2022}. 

\paragraph{Language Rotation.}
In this setup (Figure~\ref{fig:language-rotation-example}), we fix the MT model and rotate the target language at each iteration. Each sentence starts in a source language, is translated into English, then into another language, continuing cyclically through a triplet of languages (not including English). We experiment with two types of triplets: low-diversity (genealogically and geographically similar languages) and high-diversity (genealogically and geographically distant, often including low-resource languages). In a variant of this setup, we bypass English entirely and rotate translations directly among the three languages without a fixed pivot.

Regarding the selection of intermediary languages for our language rotation strategy: we started with a pool of 24 languages and used GPT-4o~\cite{openai2023gpt4} to generate initial candidate clusters based on general linguistic similarity. These initial groupings were then manually refined by the authors using typological and genealogical resources: WALS~\cite{wals2011} and Glottolog~\cite{glottolog48}. The final sets of triplets are provided in Appendix Tables~\ref{table:lowdiv} and~\ref{table:highdiv}.

We group translation chains that share the same model and language setup, compute their refined scores, and use those scores as training supervision for our evaluation models.

\subsection{Model Architecture and Training}
We adopt the \texttt{COMET} and \texttt{xCOMET} architecture~\cite{rei2020comet, rei2022cometkiwi, guerreiro2024xcomet} and train models under three standard input modes: reference-free or Quality Estimation (SRC), reference-based Regression (SRC+REF), and unified (SRC, REF, MT). The reference-free or QE mode predicts quality from source and translation only; the unified mode~\cite{wan-etal-2022-unite} supports all input combinations. Inputs are constructed by embedding the source, machine translation, and/or reference (depending on the configuration), and combining their embeddings.
All inputs are passed through a pretrained multilingual encoder: \texttt{XLM-RoBERTa-large} \cite{conneau2020xlmr} for the QE and regression models, and \texttt{InfoXLM-Large} \cite{chi2021infoxlm} for the unified variant. Layer representations from the encoder are combined using a sparsemax-weighted scalar mix~\cite{martins2016from}. The resulting [CLS] token is fed into a two-layer feedforward network to produce the final sentence-level quality score.

All models are trained using mean squared error (MSE) against our degradation-aware refined scores. We freeze encoder embeddings for the first 30\% of training and use the AdamW optimizer~\cite{loshchilov2019decoupled} with a learning rate of 1.5e-5 (1e-6 for the encoder), batch sizes of 16–32, and layerwise learning rate decay of 0.95. Training runs for up to 5 epochs with early stopping on validation loss.

\subsection{Model Variants}
We present all model variants, ours and existing automatic metrics that we use as a baseline; grouped by reference usage and supervision type, as summarized in Table~\ref{tab:model_variants}.

Alongside human-provided references, we also experiment with pseudo-references: translations from the previous iterations in a degradation chain. Models marked ``REL'' use pseudo-references, while ``HM'' models use gold references from WMT DA.

To further improve evaluation quality, we fine-tuned our best-performing models (\texttt{UM-MR-HM} and \texttt{QE-MR}) on human-annotated MQM data~\cite{freitag2021experts,freitag2021wmt21,freitag2022wmt22}, producing the \texttt{UM-FT-MQM} and \texttt{QE-FT-MQM} variants. This two-stage training strategy: pretraining followed by MQM finetuning, follows the approach introduced in \texttt{xCOMET}~\cite{rei2023xcomet}. The MQM dataset includes 148k unique examples across zh-en, en-de, and en-ru, split into training (104.8k), validation (14.5k), and test (28.9k) sets. The test set is later used to assess correlation with human judgments. 

\begin{table}[ht]
\centering
\scriptsize  
\setlength{\tabcolsep}{4pt}  
\begin{tabular}{ll}
\toprule
\textbf{Model Name} & \textbf{Description} \\
\midrule
\multicolumn{2}{l}{\textit{Baseline QE Metrics}} \\
\texttt{xCOMET-NR}     & \texttt{xCOMET} without reference input. \\
\texttt{COMETKIWI}     & QE-style \texttt{COMET} trained on WMT DA scores. \\
\multicolumn{2}{l}{\textit{Our QE Metrics}} \\
\texttt{QE-MR}         & QE model trained on model-rotated degradation data. \\
\texttt{QE-FT-MQM}        & \texttt{QE-MR} fine-tuned on MQM human annotations. \\
\texttt{QE-LR-LD}         & QE model+low-diversity language rotation (Eng. pivot). \\
\texttt{QE-LR-HD}         & QE model+ high-diversity language rotation (Eng. pivot). \\
\texttt{QE-LR-LDD}        & QE model+low-diversity direct multilingual transitions. \\
\texttt{QE-LR-HDD}        & QE model+high-diversity direct multilingual transitions. \\
\midrule
\multicolumn{2}{l}{\textit{Baseline Reference-Based and Unified Metrics}} \\
\texttt{xCOMET}        & \texttt{COMET} trained with human references. \\
\texttt{COMETDA}       & \texttt{COMET} trained directly on WMT DA scores. \\
\multicolumn{2}{l}{\textit{Our Reference-Based and Unified Metrics}} \\
\texttt{UM-MR-HM}      & Unified Metric+human reference (model-rotation). \\
\texttt{UM-MR-REL}     & Unified Metric+pseudo-reference (prior iteration). \\
\texttt{UM-FT-MQM}        & \texttt{UM-MR-HM} fine-tuned on MQM human annotations. \\
\texttt{REG-MR-HM}     & Regression model+human reference (model-rotation). \\
\texttt{REG-MR-REL}    & Regression model+pseudo-reference. \\

\bottomrule
\end{tabular}
\caption{Summary of model variants}
\label{tab:model_variants}
\end{table}

\begin{table}[htbp]
\centering
\small
\begin{tabular}{lcccc}
\toprule
\textbf{Metric} & \textbf{en-de} & \textbf{en-ru} & \textbf{zh-en} & \textbf{Average} \\
\midrule
\textit{QE Metrics} \\
\texttt{xCOMET-NR}       & 0.481 & 0.412 & 0.531 & 0.475 \\
\texttt{COMETKIWI}       & 0.371 & 0.339 & 0.335 & 0.348 \\
\midrule
\texttt{QE-LR-LD}           & 0.346 & 0.228 & 0.495 & 0.356 \\
\texttt{QE-LR-HD}           & 0.331 & 0.213 & 0.493 & 0.346 \\
\texttt{QE-LR-LDD}          & 0.375 & 0.275 & 0.474 & 0.375 \\
\texttt{QE-LR-HDD}          & 0.316 & 0.203 & 0.456 & 0.325 \\
\texttt{QE-MR}           & 0.378 & 0.297 & 0.513 & 0.396 \\
\texttt{QE-FT-MQM}         & \textbf{0.485} & \textbf{0.533} & \textbf{0.689} & \textbf{0.569} \\

\midrule
\multicolumn{5}{l}{\textit{Reference-Based and Unified Metrics}} \\
\texttt{xCOMET}          & \textbf{0.532} & 0.455 & 0.556 & 0.514 \\
\texttt{COMETDA}         & 0.434 & 0.402 & 0.394 & 0.410 \\
\midrule
\texttt{REG-MR-HM}       & 0.467 & 0.371 & 0.532 & 0.457 \\
\texttt{REG-MR-REL}      & 0.440 & 0.306 & 0.519 & 0.422 \\
\texttt{UM-MR-HM}        & 0.488 & 0.414 & 0.551 & 0.484 \\
\texttt{UM-MR-REL}       & 0.478 & 0.389 & 0.552 & 0.473 \\
\texttt{UM-FT-MQM} & 0.523 & \textbf{0.696} & \textbf{0.592} & \textbf{0.604} \\

\bottomrule
\end{tabular}
\caption{Pearson correlation with human annotated MQM scores across three language pairs for QE, reference-based, and unified metrics.
\vspace*{-1.5em}}
\label{tab:pearson_transposed_all_metrics_final}
\end{table}

These variants allow us to investigate the trade-offs between supervised, pseudo-supervised, and unsupervised evaluator training, as well as the efficacy of degradation-based supervision across different configurations.

\section{Experimental Results}
\label{sec:experimental_results}
\begin{table*}[t]
\centering
\begin{tabular}{lcccccccc}
\toprule
 & \multicolumn{2}{c}{en-de} & \multicolumn{2}{c}{en-ru} & \multicolumn{2}{c}{zh-en} & \multicolumn{2}{c}{Average} \\
\cmidrule(lr){2-3} \cmidrule(lr){4-5} \cmidrule(lr){6-7} \cmidrule(lr){8-9}
Metric & $acc_{eq}$ & SPA & $acc_{eq}$ & SPA & $acc_{eq}$ & SPA & $acc_{eq}$ & SPA \\
\midrule
\multicolumn{9}{l}{\textit{QE Metrics}} \\
\texttt{xCOMET-NR} & 0.454 & \textbf{0.932} & 0.519 & 0.763 & 0.449 & 0.561 & 0.474 & 0.752 \\
\texttt{COMETKIWI}       & 0.383 & 0.901 & 0.504 & 0.925 & 0.462 & 0.536 & 0.450 & 0.787 \\
\midrule
\texttt{QE-MR}          & 0.475 & 0.713 & 0.481 & 0.785 & 0.454 & 0.804 & 0.470 & 0.767 \\
\texttt{QE-FT-MQM}    & \textbf{0.508} & 0.919 & \textbf{0.659} & \textbf{0.931} & 0.452 & 0.635 & \textbf{0.540} & \textbf{0.828} \\
\texttt{QE-LR-LD}          & 0.432 & 0.683 & 0.450 & 0.721 & \textbf{0.491} & 0.563 & 0.458 & 0.656 \\
\texttt{QE-LR-HD}          & 0.448 & 0.700 & 0.426 & 0.764 & 0.446 & \textbf{0.812} & 0.440 & 0.759 \\
\texttt{QE-LR-LDD}         & 0.426 & 0.790 & 0.465 & 0.771 & 0.441 & 0.441 & 0.444 & 0.667 \\
\texttt{QE-LR-HDD}         & 0.475 & 0.746 & 0.411 & 0.770 & 0.433 & 0.708 & 0.440 & 0.741 \\
\midrule
\multicolumn{9}{l}{\textit{Reference-Based and Unified Metrics}} \\
\texttt{xCOMET}          & 0.552 & 0.905 & 0.597 & 0.728 & 0.462 & 0.712 & \textbf{0.537} & 0.782 \\
\texttt{COMETDA}         & \textbf{0.585} & \textbf{0.975} & 0.488 & \textbf{0.957} & 0.452 & 0.711 & 0.508 & \textbf{0.881} \\
\midrule
\texttt{REG-MR-HM}     & 0.503 & 0.906 & 0.488 & 0.867 & 0.462 & \textbf{0.827} & 0.484 & 0.867 \\
\texttt{REG-MR-REL}    & 0.541 & 0.736 & 0.450 & 0.801 & 0.472 & 0.762 & 0.488 & 0.766 \\
\texttt{UM-MR-HM}      & 0.492 & 0.940 & 0.566 & 0.844 & \textbf{0.483} & 0.531 & 0.514 & 0.772 \\
\texttt{UM-MR-REL}     & 0.497 & 0.925 & 0.597 & 0.856 & 0.470 & 0.429 & 0.521 & 0.737 \\
\texttt{UM-FT-MQM} & 0.497 & 0.902 & \textbf{0.605} & 0.812 & 0.464 & 0.630 & 0.522 & 0.781 \\
\bottomrule
\end{tabular}
\caption{Segment and system-level meta-evaluation using $acc_{eq}$ and SPA across three language pairs. }
\label{tab:spa_acc_eq}
\end{table*}

\subsection{Human Annotation Evaluation}

We compare our telephone-supervised evaluation models to existing metrics using WMT MQM human annotations. Table~\ref{tab:pearson_transposed_all_metrics_final} reports segment level Pearson correlations with human scores; results for other correlation metrics and DA annotations are provided in Appendix Tables~\ref{tab:spearman_transposed_all_metrics_final}--\ref{tab:spearman-da-wmt}.

To strengthen our segment level evaluation setup, we follow the WMT24 shared task design~\cite{freitag-etal-2024-llms} and  complement correlation with ``group-by-item'' accuracy with tie calibration $acc_{eq}$~\cite{deutsch2023ties}, which measures agreement with human segment-level preferences in pairwise comparisons.  

At the system level,  we adopt Soft Pairwise Accuracy (SPA)~\cite{thompson2024improving}. Unlike standard pairwise accuracy or Kendall’s $\tau$, SPA incorporates the statistical uncertainty of both metric and human rankings, ensuring that near-ties are handled consistently rather than penalized or rewarded arbitrarily.  

Pearson correlation is reported over the full MQM test set. In contrast, $acc_{eq}$ and SPA require source sentences to be available across all compared systems; for these evaluations, we selected the top three systems per language pair and used only the subset of sources common to them.

\begin{table*}[ht]
\centering
\small
\begin{tabular}{l@{\hspace{4pt}}l@{\hspace{2pt}}c@{\hspace{2pt}}cc|c@{\hspace{2pt}}cc|c@{\hspace{2pt}}cc}
\toprule
\textbf{Category} & \textbf{System} & \multicolumn{3}{c|}{\textbf{1 vs 2}} & \multicolumn{3}{c|}{\textbf{2 vs 3}} & \multicolumn{3}{c}{\textbf{1 vs 3}} \\
 & & \texttt{xCOMET} & \texttt{LR} & \texttt{MR} & \texttt{xCOMET} & \texttt{LR} & \texttt{MR} & \texttt{xCOMET} & \texttt{LR} & \texttt{MR} \\
\midrule
\multirow{3}{*}{\shortstack[l]{\textbf{Model} \\ \textbf{Rotation}}}

  & \texttt{Gemma-12b}     & 0.876 & 0.935 & \textbf{0.937} & 0.663 & \textbf{0.718} & 0.712 & 0.907 & 0.957 & \textbf{0.958} \\
  & \texttt{Qwen-2.5-14b}  & 0.791 & 0.850 & \textbf{0.864} & 0.640 & 0.681 & \textbf{0.682} & 0.832 & 0.888 & \textbf{0.902} \\
  & \texttt{NLLB-distilled} & 0.677 & \textbf{0.753} & 0.745 & 0.599 & \textbf{0.644} & 0.635 & 0.723 & \textbf{0.806} & 0.798 \\
\midrule
\multirow{3}{*}{\shortstack[l]{\textbf{Commercial} \\ \textbf{Models MR}}}

  & \texttt{GPT-4o}         & 0.955 & \textbf{0.973} & \textbf{0.973} & 0.628 & \textbf{0.706} & 0.669 & 0.966 & \textbf{0.983} & 0.981 \\
  & \texttt{GPT-4o-mini}    & 0.942 & \textbf{0.965} & 0.963 & 0.641 & \textbf{0.703} & 0.678 & 0.958 & \textbf{0.976} & 0.974 \\
  & \texttt{NLLB-distilled} & 0.879 &\textbf{ 0.931} & 0.917 & 0.616 & \textbf{0.685} & 0.649 & 0.905 & \textbf{0.950} & 0.938 \\
\midrule
\multirow{3}{*}{\shortstack[l]{\textbf{Language} \\ \textbf{Rotation}}} 
  & First Language   & 0.732 & \textbf{0.865} & 0.830 & 0.737 & \textbf{0.847} & 0.797 & 0.852 & \textbf{0.955} & 0.927 \\
  & Second Language  & 0.727 & \textbf{0.839} & 0.783 & 0.688 & \textbf{0.774} & 0.727 & 0.820 & \textbf{0.916} & 0.872 \\
  & Third Language   & 0.759 & \textbf{0.845} & 0.801 & 0.698 & \textbf{0.769} & 0.738 & 0.839 & \textbf{0.913} & 0.885 \\
\bottomrule
\end{tabular}
\caption{
AUC scores for detecting translation quality degradation across three round pairs (1 vs 2, 2 vs 3, 1 vs 3), comparing \texttt{xCOMET}, language-rotation (\texttt{LR}), and model-rotation (\texttt{MR}) variants across different setups.   Higher AUC indicates better discrimination between earlier and later degraded outputs. 
}

\label{tab:auc_combined_all_reordered}
\end{table*}

\subsubsection{Segment-Level Meta-Evaluation}

\paragraph{Reference-Free Metrics.}  
From Pearson correlation data on Table~\ref{tab:pearson_transposed_all_metrics_final}, model rotation proves the most effective degradation strategy: \texttt{QE-MR} (0.396) outperforms all language-rotated variants, with lower-diversity setups (e.g., \texttt{QE-LR-LDD} at 0.375) outperforms high-diversity ones, hinting that excessive linguistic variation may dilute the training signal. Despite being trained entirely without human labels, \texttt{QE-MR} performs competitively and surpasses \texttt{COMETKIWI} (0.348). Fine-tuning on MQM further boosts performance: \texttt{QE-FT-MQM} leads all reference-free models with 0.569, outperforming \texttt{xCOMET-NR} (0.475). These results suggest that degradation-based pretraining provides a solid foundation, strong on its own, and even stronger when combined with human-labeled supervision. The same pattern is consistently reflected under $acc_{eq}$ results in Table~\ref{tab:spa_acc_eq}.

\paragraph{Reference-Based Metrics.}  
\texttt{UM-FT-MQM} leads with the highest average Pearson correlation (0.604), showing strong gains on en--ru (0.696) and zh--en (0.592), and surpassing the established \texttt{xCOMET} (0.514). Our degradation-trained models, especially \texttt{UM-MR-HM} (0.484) and \texttt{UM-MR-REL} (0.473), closely trail behind, showing that pseudo-references can be nearly as effective as gold ones. Regression variants like \texttt{REG-MR-HM} (0.457) and \texttt{REG-MR-REL} (0.422) remain competitive, while \texttt{COMETDA} (0.410) trails slightly, likely due to its training with DA-specific data.

When evaluated with $acc_{eq}$ (Table~\ref{tab:spa_acc_eq}), \texttt{xCOMET} (0.537) achieves the strongest agreement with human segment-level preferences, followed closely by \texttt{UM-FT-MQM} (0.522) and \texttt{UM-MR-REL} (0.521). \texttt{UM-MR-HM} (0.514) remains competitive. Interestingly, \texttt{COMETDA}, which lagged on Pearson correlation, performs reasonably well under $acc_{eq}$ (0.508), indicating that while it struggles with fine-grained score alignment, it remains effective at identifying the better translation in pairwise comparisons.

\subsubsection{System-Level Meta-Evaluation}
\paragraph{Reference-Free Metrics.}  
SPA results in Table~\ref{tab:spa_acc_eq} reveal that reference-free models generally lag behind their reference-based counterparts in system-level ranking. 
Among them, \texttt{QE-FT-MQM} stands out with the highest SPA (0.828), showing that fine-tuning on MQM not only improves segment-level correlation but also enhances robustness when comparing systems. Interestingly, high diversity language-rotation variants such as \texttt{QE-LR-HD} (0.759) and \texttt{QE-LR-HDD} (0.741) achieve relatively strong SPA despite weaker segment level scores.

\paragraph{Reference-Based Metrics.}  
Here SPA highlights a different set of winners than Pearson correlation. The baseline \texttt{COMETDA} achieves the highest score (0.881), while \texttt{xCOMET} reaches 0.782. Among our models, regression variants perform best, with \texttt{REG-MR-HM} at 0.867. Unified metrics such as \texttt{UM-FT-MQM} (0.781) and \texttt{UM-MR-HM} (0.772) remain competitive but do not surpass the baselines, reflecting their strength in segment-level evaluation rather than system-level ranking.

\subsection{Paired Generation Evaluation}
\label{subsec:paired_generation}
A core assumption of our framework is that earlier iterations in a translation chain should retain higher quality than later ones. To test whether evaluation metrics capture this implicit ordering, we conduct a paired generation test: given two outputs (e.g., from round 1 and 3), can the metric correctly identify the better one? We treat this as a binary discrimination task and report AUC scores for all pairwise comparisons among rounds: 1 vs 2, 2 vs 3, and 1 vs 3, using outputs from model-rotated chains, language-rotated chains, and model-rotated chains including commercial systems like \texttt{GPT-4o}. We compare three representative metrics: \texttt{xCOMET}, our language-rotation model \texttt{QE-LR-LD} (\texttt{LR}), and its model-rotation counterpart \texttt{QE-MR} (\texttt{MR}). 
\begin{table*}[ht]
\centering
\begin{tabular}{lccc|ccc|ccc}
\toprule
\textbf{Model} & \multicolumn{3}{c|}{\textbf{1 vs 2}} & \multicolumn{3}{c|}{\textbf{2 vs 3}} & \multicolumn{3}{c}{\textbf{1 vs 3}} \\
 & \texttt{UM} & \texttt{IT} & \texttt{UNMD} & \texttt{UM} & \texttt{IT} & \texttt{UNMD} & \texttt{UM} & \texttt{IT} & \texttt{UNMD} \\
\midrule
\texttt{Gemma-12b}       & 0.944 & \textbf{0.989} & 0.913 & 0.754 & \textbf{0.789} & 0.692 & 0.962 & \textbf{0.993} & 0.937 \\
\texttt{Qwen-2.5-14b}    & 0.882 & \textbf{0.943} & 0.825 & 0.719 & \textbf{0.739} & 0.665 & 0.913 & \textbf{0.961} & 0.864 \\
\texttt{NLLB-distilled}  & 0.786 & \textbf{0.834} & 0.710 & 0.675 & \textbf{0.679} & 0.614 & 0.835 & \textbf{0.883} & 0.757 \\
\bottomrule
\end{tabular}
\caption{
AUC scores for degradation detection across round pairs (1 vs 2, 2 vs 3, 1 vs 3), comparing \texttt{UM-MR-HM} (\texttt{UM}), iteration-averaged (\texttt{IT}), and unmodified \texttt{xCOMET}-trained (\texttt{UNMD}) models on three MT systems. 
\vspace*{-1em}
}
\label{tab:auc_ablation_variants}
\end{table*}

As shown in Table~\ref{tab:auc_combined_all_reordered}, distinguishing between later iterations, especially 2 vs 3 is consistently the most difficult. AUC scores drop sharply here, with \texttt{xCOMET} reaching just 0.599 on \texttt{NLLB} and 0.628 on \texttt{GPT-4o}. In contrast, our degradation-trained models show greater resilience: \texttt{MR} improves to 0.635 and 0.669 on the same systems, while \texttt{LR} does even better, achieving 0.644 and 0.706.

Surprisingly, \texttt{LR} despite scoring lower on human correlation, outperforms \texttt{MR} in 8 out of 9 commercial comparisons and performs comparably in model-rotation settings. Its advantage is most pronounced in language-rotated chains, where it achieves up to 0.847 (2 vs 3) and 0.955 (1 vs 3), suggesting a stronger capacity to track structural shifts across languages. On high-quality outputs from \texttt{GPT-4o}, both \texttt{LR} and \texttt{MR} maintain near-ceiling AUCs (up to 0.983), consistently outperforming \texttt{xCOMET} across all comparisons.

\paragraph{Training Signal Variants.}
\label{sec:ablation}

We compare three training strategies on the paired generation test (Table~\ref{tab:auc_ablation_variants}). \texttt{UM-MR-HM} uses our refined scores. \texttt{IT} ignores sentence-level detail, assigning a single average score per iteration. \texttt{UNMD} is trained directly on \texttt{xCOMET} scores, without degradation awareness.
 \texttt{IT} performs best across all systems; for example, scoring 0.961 (1 vs 3) and 0.739 (2 vs 3) on \texttt{Qwen-2.5}, outperforming \texttt{UM-MR-HM} at 0.913 and 0.719. Since \texttt{IT} only knows which iteration a sentence came from, it learns to align position with quality, an advantage in relative ranking. \texttt{UM-MR-HM} still performs well, showing that sentence-level signals help capture more detail even if they aren’t essential for ranking which output is better. \texttt{UNMD} lags behind, confirming the value of degradation-based supervision.

\section{Conclusion}
\label{sec:discussion}

We have demonstrated that synthetic translation data produced using language and model rotation strategies can be used to train state-of-the-art translation evaluation models.   Future work should focus on better methods to score the training data as a function of generation and text complexity, and applying this methodology to more challenging long-form tasks.

\newpage

\section*{Limitations}
\label{sec:limitations}
While promising, our method has several limitations. First, the degradation trajectories we induce may not fully reflect real-world translation errors. They tend to emphasize lexical and syntactic shifts, potentially underrepresenting issues such as discourse inconsistency or cultural misalignment. Additionally, the quality of our pseudo-labels ultimately depends on the reliability of the scoring anchor (\texttt{xCOMET}); any bias or blind spots in this metric propagate to the training signal.
Finally, we assume access to a diverse set of MT systems and high-coverage multilingual translation capabilities. In truly low-resource settings with poor model availability, both rotation strategies may be difficult to execute meaningfully, constraining the generality of our approach.

\appendix
\onecolumn

\section{Appendix}
\label{appendix}

\begin{table}[h]
\centering
\begin{tabular}{ll}
\toprule
Source Language & Low-Diversity Triplet \\
\midrule
Bengali & Bengali, Hindi, Gujarati \\
Central Khmer & Central Khmer, Chinese, Japanese \\
Chinese & Chinese, Central Khmer, Japanese \\
Czech & Czech, Polish, Russian \\
Estonian & Estonian, Finnish, Latvian \\
Finnish & Finnish, Estonian, Latvian \\
French & French, German, Polish \\
German & German, French, Polish \\
Gujarati & Gujarati, Hindi, Bengali \\
Hausa & Hausa, Zulu, Xhosa \\
Hindi & Hindi, Gujarati, Bengali \\
Icelandic & Icelandic, German, French \\
Japanese & Japanese, Chinese, Central Khmer \\
Kazakh & Kazakh, Russian, Ukrainian \\
Latvian & Latvian, Lithuanian, Estonian \\
Lithuanian & Lithuanian, Latvian, Estonian \\
Pashto & Pashto, Hindi, Kazakh \\
Polish & Polish, Czech, German \\
Russian & Russian, Ukrainian, Kazakh \\
Tamil & Tamil, Hindi, Gujarati \\
Turkish & Turkish, Kazakh, Russian \\
Ukrainian & Ukrainian, Russian, Kazakh \\
Xhosa & Xhosa, Zulu, Hausa \\
Zulu & Zulu, Xhosa, Hausa \\
\bottomrule
\end{tabular}
\caption{Low-diversity triplets used in our experiments.}
\label{table:lowdiv}
\end{table}

\begin{table}[h]
\centering
\begin{tabular}{ll}
\toprule
Source Language & High-Diversity Triplet \\
\midrule
Bengali & Bengali, Russian, Hausa \\
Central Khmer & Central Khmer, Turkish, Finnish \\
Chinese & Chinese, Zulu, Lithuanian \\
Czech & Czech, Japanese, Pashto \\
Estonian & Estonian, Hindi, Xhosa \\
Finnish & Finnish, Tamil, Kazakh \\
French & French, Gujarati, Central Khmer \\
German & German, Pashto, Japanese \\
Gujarati & Gujarati, Ukrainian, Zulu \\
Hausa & Hausa, Chinese, Latvian \\
Hindi & Hindi, Estonian, Kazakh \\
Icelandic & Icelandic, Bengali, Japanese \\
Japanese & Japanese, Hausa, Latvian \\
Kazakh & Kazakh, Xhosa, Polish \\
Latvian & Latvian, Tamil, Japanese \\
Lithuanian & Lithuanian, Hindi, Zulu \\
Pashto & Pashto, French, Chinese \\
Polish & Polish, Central Khmer, Gujarati \\
Russian & Russian, Estonian, Zulu \\
Tamil & Tamil, German, Ukrainian \\
Turkish & Turkish, Japanese, Lithuanian \\
Ukrainian & Ukrainian, Hausa, Chinese \\
Xhosa & Xhosa, Russian, Hindi \\
Zulu & Zulu, French, Kazakh \\

\bottomrule
\end{tabular}
\caption{High-diversity triplets used in our experiments.}
\label{table:highdiv}
\end{table}

\begin{table}[htbp]
\centering
\small
\begin{tabular}{lcccc}
\toprule
\textbf{Metric} & \textbf{en-de} & \textbf{en-ru} & \textbf{zh-en} & \textbf{Average} \\
\midrule
\textit{QE Metrics} \\
\texttt{xCOMET-NR}       & \textbf{0.410} & 0.468 & 0.507 & 0.462 \\
\texttt{COMETKIWI}       & 0.306 & 0.391 & 0.372 & 0.356 \\
\midrule
\texttt{QE-LR-LD}           & 0.315 & 0.312 & 0.472 & 0.366 \\
\texttt{QE-LR-HD}           & 0.314 & 0.284 & 0.478 & 0.359 \\
\texttt{QE-LR-LDD}          & 0.342 & 0.342 & 0.465 & 0.383 \\
\texttt{QE-LR-HDD}         & 0.326 & 0.286 & 0.480 & 0.364 \\
\texttt{QE-MR}              & 0.328 & 0.366 & 0.491 & 0.395 \\
\texttt{QE-FT-MQM}          & 0.368 & \textbf{0.474} & \textbf{0.558} & \textbf{0.467} \\
\midrule
\multicolumn{5}{l}{\textit{Reference-Based and Unified Metrics}} \\
\texttt{xCOMET}             & \textbf{0.476} & 0.513 & 0.528 & 0.506 \\
\texttt{COMETDA}            & 0.419 & 0.433 & 0.439 & 0.430 \\
\midrule
\texttt{REG-MR-HM }         & 0.412 & 0.407 & 0.507 & 0.442 \\
\texttt{REG-MR-REL}         & 0.401 & 0.391 & 0.496 & 0.429 \\
\texttt{UM-MR-HM}           & 0.436 & 0.442 & 0.519 & 0.466 \\
\texttt{UM-MR-REL}          & 0.422 & 0.418 & 0.514 & 0.451 \\
\texttt{UM-FT-MQM}          & 0.432 & \textbf{0.581} & \textbf{0.536} & \textbf{0.516} \\
\bottomrule
\end{tabular}
\caption{Spearman correlation with human annotated MQM scores across three language pairs for QE, reference-based, and unified metrics.}
\label{tab:spearman_transposed_all_metrics_final}
\end{table}

\begin{table}[htbp]
\centering
\small
\begin{tabular}{lcccc}
\toprule
\textbf{Metric} & \textbf{en-de} & \textbf{en-ru} & \textbf{zh-en} & \textbf{Average} \\
\midrule
\textit{QE Metrics} \\
\texttt{xCOMET-NR}       & \textbf{0.320} & \textbf{0.357} & 0.382 & 0.353 \\
\texttt{COMETKIWI}       & 0.229 & 0.288 & 0.271 & 0.263 \\
\midrule
\texttt{QE-LR-LD}           & 0.235 & 0.228 & 0.350 & 0.271 \\
\texttt{QE-LR-HD}           & 0.234 & 0.206 & 0.353 & 0.264 \\
\texttt{QE-LR-LD}D          & 0.257 & 0.251 & 0.343 & 0.284 \\
\texttt{QE-LR-HD}D          & 0.243 & 0.208 & 0.354 & 0.268 \\
\texttt{QE-MR}           & 0.246 & 0.269 & 0.364 & 0.293 \\
\texttt{QE-FT-MQM}       & 0.284 & 0.354 & \textbf{0.431} & \textbf{0.356} \\
\midrule
\multicolumn{5}{l}{\textit{Reference-Based and Unified Metrics}} \\
\texttt{xCOMET}          & \textbf{0.372} & 0.389 & 0.399 & 0.387 \\
\texttt{COMETDA}         & 0.319 & 0.321 & 0.323 & 0.321 \\
\midrule
\texttt{REG-MR-HM }      & 0.312 & 0.301 & 0.375 & 0.331 \\
\texttt{REG-MR-REL}      & 0.310 & 0.291 & 0.375 & 0.325 \\
\texttt{UM-MR-HM}        & 0.338 & 0.332 & 0.391 & 0.354 \\
\texttt{UM-MR-REL}       & 0.320 & 0.309 & 0.385 & 0.338 \\
\texttt{UM-FT-MQM}          & 0.329 & \textbf{0.448} & \textbf{0.401} & \textbf{0.393} \\
\bottomrule
\end{tabular}
\caption{Kendall’s Tau correlation with human annotated MQM scores across three language pairs for QE, reference-based, and unified metrics.}
\label{tab:tau_transposed_all_metrics_final}
\end{table}

\begin{table*}[ht]
\centering
\small
\begin{tabular}{lccccccccc|c}
\toprule
\textbf{Metric} & \textbf{en-cs} & \textbf{en-de} & \textbf{en-fi} & \textbf{en-ru} & \textbf{en-tr} & \textbf{en-zh} & \textbf{cs-en} & \textbf{fi-en} & \textbf{de-en} & \textbf{Average} \\
\midrule
\multicolumn{11}{l}{\textit{Reference-Free (QE) Metrics}} \\
\midrule
\texttt{xCOMET-NR}     & 0.658 & 0.570 & 0.725 & 0.564 & 0.631 & 0.343 & 0.171 & 0.474 & 0.257 & 0.488 \\
\texttt{COMETKIWI}         & \textbf{0.741} & \textbf{0.593} & \textbf{0.770} & \textbf{0.617} & \textbf{0.692} & \textbf{0.378} & \textbf{0.246} & \textbf{0.515} & \textbf{0.315} & \textbf{0.541} \\
\texttt{QE-MR}             & 0.521 & 0.411 & 0.633 & 0.474 & 0.582 & 0.267 & 0.136 & 0.417 & 0.202 & 0.404 \\
\texttt{LR-LD}             & 0.470 & 0.346 & 0.591 & 0.407 & 0.509 & 0.204 & 0.102 & 0.389 & 0.177 & 0.355 \\
\texttt{LR-HD}             & 0.425 & 0.286 & 0.531 & 0.351 & 0.478 & 0.201 & 0.065 & 0.396 & 0.148 & 0.331 \\
\texttt{LR-LDD}            & 0.502 & 0.382 & 0.602 & 0.452 & 0.516 & 0.207 & 0.122 & 0.385 & 0.192 & 0.384 \\
\texttt{LR-HDD}            & 0.421 & 0.262 & 0.494 & 0.327 & 0.403 & 0.168 & 0.068 & 0.339 & 0.109 & 0.288 \\
\midrule
\multicolumn{11}{l}{\textit{Reference-Based and Unified Metrics (\texttt{xCOMET} and Variants)}} \\
\midrule
\texttt{xCOMET}            & 0.715 & \textbf{0.605} & 0.744 & 0.604 & 0.694 & 0.401 & 0.222 & 0.509 & 0.308 & 0.533 \\
\texttt{COMETDA}          & \textbf{0.745} & 0.598 & \textbf{0.764} & \textbf{0.605} & \textbf{0.738} & \textbf{0.449} & \textbf{0.236} & \textbf{0.547} & \textbf{0.349} & \textbf{0.559} \\
\texttt{\texttt{UM-MR-HM}}          & 0.653 & 0.530 & 0.712 & 0.567 & 0.667 & 0.374 & 0.200 & 0.493 & 0.282 & 0.497 \\
\texttt{\texttt{UM-MR-REL}}         & 0.630 & 0.533 & 0.703 & 0.562 & 0.658 & 0.350 & 0.192 & 0.494 & 0.278 & 0.489 \\
\texttt{REG-MR-HM}         & 0.605 & 0.501 & 0.687 & 0.523 & 0.601 & 0.299 & 0.163 & 0.464 & 0.257 & 0.456 \\
\texttt{\texttt{REG-MR-REL}}        & 0.612 & 0.518 & 0.666 & 0.504 & 0.614 & 0.309 & 0.151 & 0.438 & 0.236 & 0.450 \\
\texttt{UM-FT-MQM}            & 0.612 & 0.518 & 0.666 & 0.523 & 0.614 & 0.314 & 0.160 & 0.432 & 0.222 & 0.446 \\
\bottomrule
\end{tabular}
\caption{
Spearman correlation with DA human annotations across selected language pairs. The average is for all 36 language pairs. While DA-trained models (e.g., \texttt{COMETDA}, \texttt{COMETKIWI}) achieve the highest scores—as expected due to fine-tuning directly on DA data. Our models (e.g., \texttt{UM-MR-HM}, \texttt{REG-MR-HM}) perform competitively despite not being trained on this supervision. 
}
\label{tab:spearman-da-wmt}
\end{table*}

\begin{figure*}[ht]
    \centering
    \begin{subfigure}[t]{0.32\textwidth}
        \includegraphics[width=\linewidth]{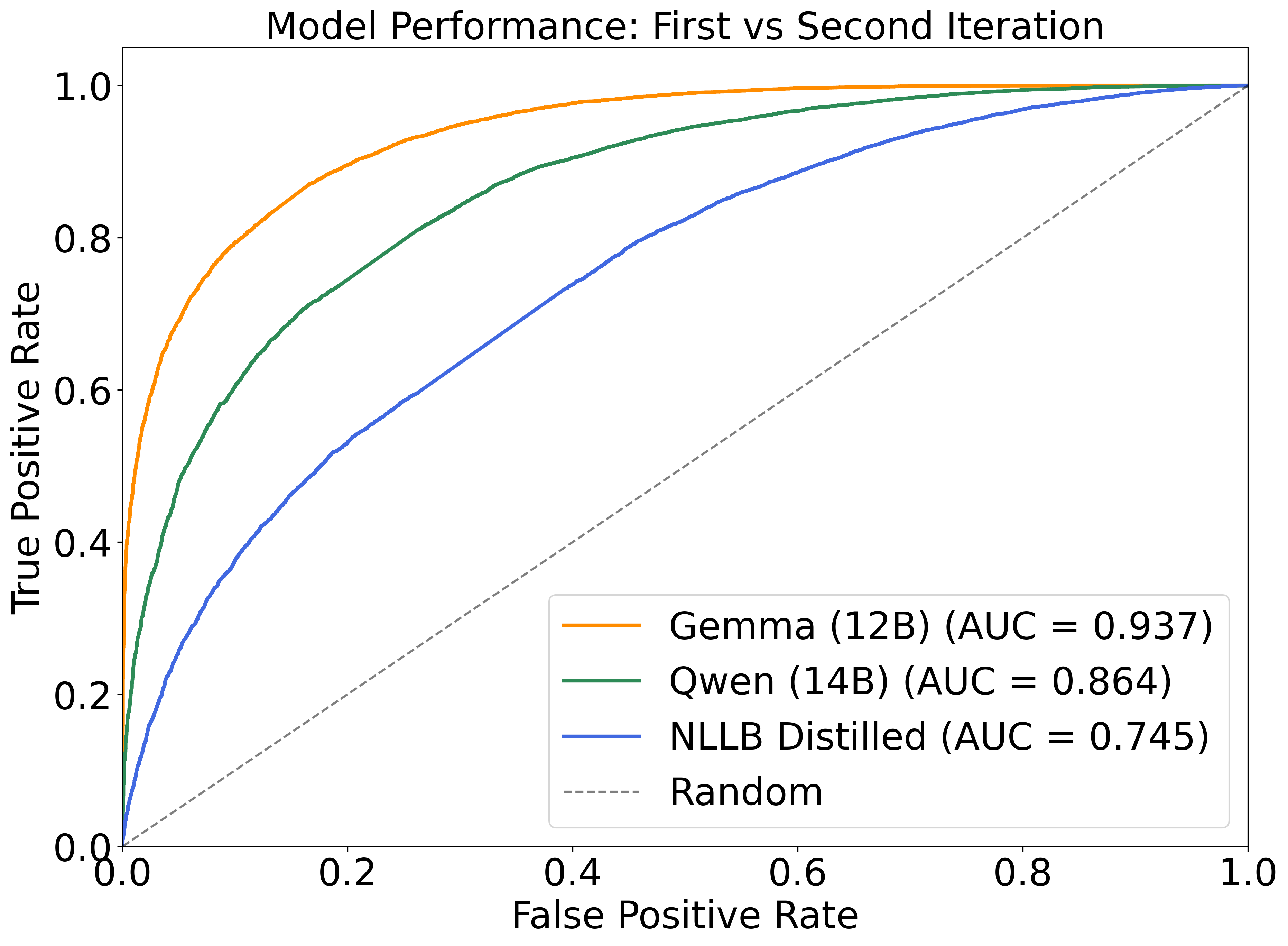}
        \caption{1 vs 2}
    \end{subfigure}
    \hfill
    \begin{subfigure}[t]{0.32\textwidth}
        \includegraphics[width=\linewidth]{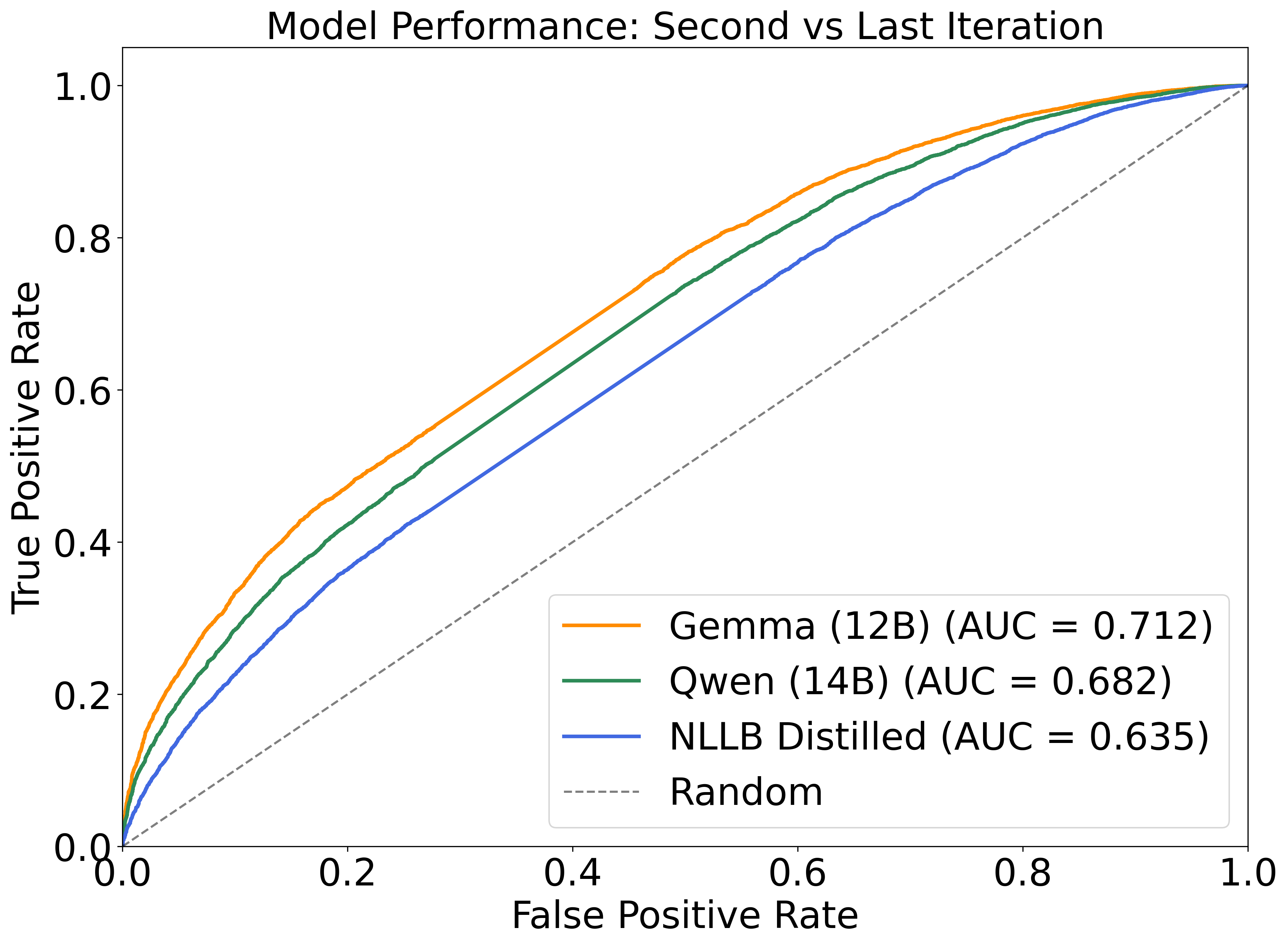}
        \caption{2 vs 3}
    \end{subfigure}
    \hfill
    \begin{subfigure}[t]{0.32\textwidth}
        \includegraphics[width=\linewidth]{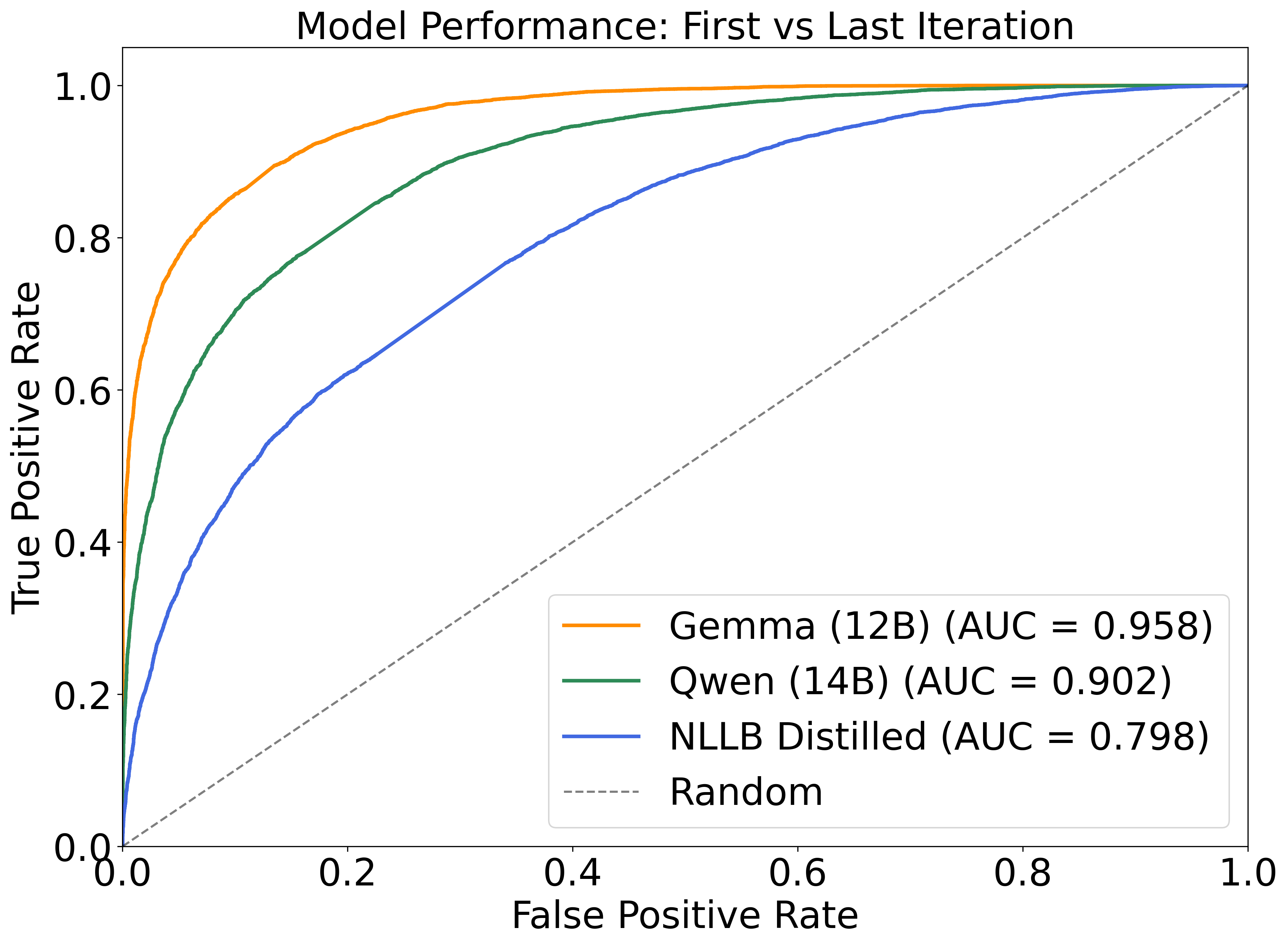}
        \caption{1 vs 3}
    \end{subfigure}
    \caption{
AUC curves for detecting translation degradation using \texttt{QE-MR} with outputs from three MT systems in the model rotation chain: \texttt{Gemma-12B}, \texttt{Qwen-14B}, and \texttt{NLLB-distilled}. Each subplot compares model predictions between different iteration pairs—(a) 1 vs 2, (b) 2 vs 3, and (c) 1 vs 3. Higher AUC indicates better discrimination between translation quality shifts over iterations.
}
    \label{fig:auc_plots}
\end{figure*}


\begin{thebibliography}{42}
\providecommand{\natexlab}[1]{#1}

\bibitem[{Agrawal et~al.(2023)Agrawal, Farinhas, Rei, and Martins}]{agrawal2023can}
Sweta Agrawal, Ant{\^o}nio Farinhas, Ricardo Rei, and Andr{\'e} F.~T. Martins. 2023.
\newblock Can automatic metrics assess high-quality translations?
\newblock In \emph{Proceedings of the 61st Annual Meeting of the Association for Computational Linguistics (ACL)}. Association for Computational Linguistics.

\bibitem[{Anil et~al.(2024)Anil, Huang, Dai, Narang, Lu, Chowdhery, Dohan, Siddhant, Freitas, Levskaya, and et~al.}]{anil2024gemma}
Rohan Anil, Yanping Huang, Andrew~M Dai, Sharan Narang, Yifeng Lu, Aakanksha Chowdhery, David Dohan, Aditya Siddhant, Daniel~De Freitas, Anselm Levskaya, and et~al. 2024.
\newblock \href {https://arxiv.org/abs/2403.07672} {Gemma: Open models based on gemini research and technology}.
\newblock \emph{Preprint}, arXiv:2403.07672.

\bibitem[{Banerjee and Lavie(2005)}]{banerjee2005meteor}
Satanjeev Banerjee and Alon Lavie. 2005.
\newblock Meteor: An automatic metric for mt evaluation with improved correlation with human judgments.
\newblock In \emph{ACL Workshop}, pages 65--72.

\bibitem[{Beinborn and Choenni(2020)}]{beinbornchoenni2020semantic}
Lisa Beinborn and Rochelle Choenni. 2020.
\newblock \href {https://doi.org/10.1162/coli_a_00382} {Semantic drift in multilingual representations}.
\newblock \emph{Computational Linguistics}, 46(3):571--603.

\bibitem[{Chi et~al.(2021)Chi, Dong, Wei, Wang, Mao, Huang, and Zhou}]{chi2021infoxlm}
Zewen Chi, Li~Dong, Furu Wei, Wenhui Wang, Xian-Ling Mao, Heyan Huang, and Ming Zhou. 2021.
\newblock \href {https://doi.org/10.18653/v1/2021.naacl-main.280} {Infoxlm: An information-theoretic framework for cross-lingual language model pre-training}.
\newblock In \emph{Proceedings of the 2021 Conference of the North American Chapter of the Association for Computational Linguistics: Human Language Technologies}, pages 3576--3588. Association for Computational Linguistics.

\bibitem[{Conneau et~al.(2020)Conneau, Khandelwal, Goyal, Chaudhary, Wenzek, Guzm{\'a}n, Grave, Ott, Zettlemoyer, and Stoyanov}]{conneau2020xlmr}
Alexis Conneau, Kartikay Khandelwal, Naman Goyal, Vishrav Chaudhary, Guillaume Wenzek, Francisco Guzm{\'a}n, Edouard Grave, Myle Ott, Luke Zettlemoyer, and Veselin Stoyanov. 2020.
\newblock \href {https://doi.org/10.18653/v1/2020.acl-main.747} {Unsupervised cross-lingual representation learning at scale}.
\newblock In \emph{Proceedings of the 58th Annual Meeting of the Association for Computational Linguistics}, pages 8440--8451. Association for Computational Linguistics.

\bibitem[{Costa-jussà et~al.(2022)Costa-jussà, Cross, Çelebi, Elbayad, Heafield, Heffernan et~al.}]{nllb2022}
Marta~R. Costa-jussà, James Cross, Onur Çelebi, Maha Elbayad, Kenneth Heafield, Kevin Heffernan, and 1 others. 2022.
\newblock \href {https://aclanthology.org/2022.emnlp-main.17} {No language left behind: Scaling human-centered machine translation}.
\newblock In \emph{Proceedings of the 2022 Conference on Empirical Methods in Natural Language Processing}.

\bibitem[{Cui et~al.(2024)Cui, Bai, Wang, Lin, Zhuang, Li, and et~al.}]{cui2024qwen2}
Yuxiao Cui, Qingyang Bai, Weizhi Wang, Junyang Lin, Bohan Zhuang, Lei Li, and et~al. 2024.
\newblock \href {https://arxiv.org/abs/2401.04088} {Qwen2: Scaling up language models in chinese and english}.
\newblock \emph{Preprint}, arXiv:2401.04088.

\bibitem[{Deutsch et~al.(2023)Deutsch, Foster, and Freitag}]{deutsch2023ties}
Daniel Deutsch, George Foster, and Markus Freitag. 2023.
\newblock Ties matter: Meta-evaluating modern metrics with pairwise accuracy and tie calibration.
\newblock \emph{arXiv preprint arXiv:2305.14324}.

\bibitem[{Dryer and Haspelmath(2011)}]{wals2011}
Matthew~S. Dryer and Martin Haspelmath, editors. 2011.
\newblock \href {https://wals.info} {\emph{The World Atlas of Language Structures Online}}.
\newblock Max Planck Institute for Evolutionary Anthropology, Leipzig.

\bibitem[{Etchegoyhen and Ponce(2023)}]{etchegoyhen2023learning}
Thierry Etchegoyhen and David Ponce. 2023.
\newblock \href {https://aclanthology.org/2023.mtsummit-research.8} {Learning from past mistakes: Quality estimation from monolingual corpora and machine translation learning stages}.
\newblock In \emph{Proceedings of Machine Translation Summit XIX, Vol 1: Research Track}, pages 84--98, Macau, China. Asia-Pacific Association for Machine Translation.

\bibitem[{Fan et~al.(2021)Fan, Bhosale, Schwenk, Ma, El-Kishky, Goyal, Baines, Celebi, Wenzek, Chaudhary, Goyal, Birch, Liptchinsky, Edunov, Grave, Auli, and Joulin}]{fan2021beyond}
Angela Fan, Shruti Bhosale, Holger Schwenk, Zhiyi Ma, Ahmed El-Kishky, Siddharth Goyal, Mandeep Baines, Onur Celebi, Guillaume Wenzek, Vishrav Chaudhary, Naman Goyal, Tom Birch, Vitaliy Liptchinsky, Sergey Edunov, Edouard Grave, Michael Auli, and Armand Joulin. 2021.
\newblock \href {https://www.jmlr.org/papers/volume22/20-1307/20-1307.pdf} {Beyond english-centric multilingual machine translation}.
\newblock \emph{Journal of Machine Learning Research}, 22:1--48.

\bibitem[{Fomicheva et~al.(2020)Fomicheva, Shterionov, Guzm{\'a}n, Specia, and Martins}]{fomicheva2020unsupervised}
Marina Fomicheva, Dimitar Shterionov, Francisco Guzm{\'a}n, Lucia Specia, and Andr{\'e}~FT Martins. 2020.
\newblock Unsupervised quality estimation for neural machine translation.
\newblock In \emph{Proceedings of the 2020 Conference on Empirical Methods in Natural Language Processing (EMNLP)}.

\bibitem[{Freitag et~al.(2022{\natexlab{a}})Freitag, Al-Onaizan, Ma et~al.}]{freitag2022high}
Markus Freitag, Yaser Al-Onaizan, Shuo~Sun Ma, and 1 others. 2022{\natexlab{a}}.
\newblock High-quality low-resource machine translation: A new benchmark.
\newblock In \emph{Findings of EMNLP}.

\bibitem[{Freitag et~al.(2021{\natexlab{a}})Freitag, Foster, Grangier, Ratnakar, Tan, and Macherey}]{freitag2021experts}
Markus Freitag, George Foster, David Grangier, Viresh Ratnakar, Qijun Tan, and Wolfgang Macherey. 2021{\natexlab{a}}.
\newblock \href {https://doi.org/10.1162/tacl_a_00437} {Experts, errors, and context: A large-scale study of human evaluation for machine translation}.
\newblock \emph{Transactions of the Association for Computational Linguistics}, 9:1460--1474.

\bibitem[{Freitag et~al.(2024)Freitag, Mathur, Deutsch, Lo, Avramidis, Rei, Thompson, Blain, Kocmi, Wang, Adelani, Buchicchio, Zerva, and Lavie}]{freitag-etal-2024-llms}
Markus Freitag, Nitika Mathur, Daniel Deutsch, Chi-Kiu Lo, Eleftherios Avramidis, Ricardo Rei, Brian Thompson, Frederic Blain, Tom Kocmi, Jiayi Wang, David~Ifeoluwa Adelani, Marianna Buchicchio, Chrysoula Zerva, and Alon Lavie. 2024.
\newblock \href {https://doi.org/10.18653/v1/2024.wmt-1.2} {Are {LLM}s breaking {MT} metrics? results of the {WMT}24 metrics shared task}.
\newblock In \emph{Proceedings of the Ninth Conference on Machine Translation}, pages 47--81, Miami, Florida, USA. Association for Computational Linguistics.

\bibitem[{Freitag et~al.(2021{\natexlab{b}})Freitag, Mathur, Bojar et~al.}]{freitag2021results}
Markus Freitag, Prashant Mathur, Ondřej Bojar, and 1 others. 2021{\natexlab{b}}.
\newblock Results of the wmt21 metrics shared task: Evaluating metrics with expert-based human evaluations on ted and news tasks.
\newblock In \emph{Proceedings of the Sixth Conference on Machine Translation}, pages 733--774.

\bibitem[{Freitag et~al.(2022{\natexlab{b}})Freitag, Rei, Mathur, Lo, Stewart, Avramidis, Kocmi, Foster, Lavie, and Martins}]{freitag2022wmt22}
Markus Freitag, Ricardo Rei, Nitika Mathur, Chi-kiu Lo, Craig Stewart, Eleftherios Avramidis, Tom Kocmi, George Foster, Alon Lavie, and André F.~T. Martins. 2022{\natexlab{b}}.
\newblock \href {https://aclanthology.org/2022.wmt-1.2} {Results of wmt22 metrics shared task: Stop using bleu – neural metrics are better and more robust}.
\newblock In \emph{Proceedings of the Seventh Conference on Machine Translation}, pages 46--68, Abu Dhabi. Association for Computational Linguistics.

\bibitem[{Freitag et~al.(2021{\natexlab{c}})Freitag, Rei, Mathur, Lo, Stewart, Foster, Lavie, and Bojar}]{freitag2021wmt21}
Markus Freitag, Ricardo Rei, Nitika Mathur, Chi-kiu Lo, Craig Stewart, George Foster, Alon Lavie, and Ondřej Bojar. 2021{\natexlab{c}}.
\newblock \href {https://aclanthology.org/2021.wmt-1.73} {Results of the wmt21 metrics shared task: Evaluating metrics with expert-based human evaluations on ted and news domain}.
\newblock In \emph{Proceedings of the Sixth Conference on Machine Translation}, pages 733--774, Online. Association for Computational Linguistics.

\bibitem[{Glushkova et~al.(2021)Glushkova, Rei, Farinha, and Specia}]{glushkova2021uncertainty}
Taisiya Glushkova, Ricardo Rei, Ana Farinha, and Lucia Specia. 2021.
\newblock Uncertainty-aware comet: A confidence estimation framework for mt evaluation.
\newblock In \emph{WMT}, pages 1026--1035.

\bibitem[{Guerreiro et~al.(2024)Guerreiro, Rei, Stigt, Coheur, Colombo, and Martins}]{guerreiro2024xcomet}
Nuno~M Guerreiro, Ricardo Rei, Daan~van Stigt, Luisa Coheur, Pierre Colombo, and Andr{\'e}~FT Martins. 2024.
\newblock xcomet: Transparent machine translation evaluation through fine-grained error detection.
\newblock \emph{Transactions of the Association for Computational Linguistics}, 12:979--995.

\bibitem[{Hammarstr{\"o}m et~al.(2023)Hammarstr{\"o}m, Forkel, Haspelmath, and Bank}]{glottolog48}
Harald Hammarstr{\"o}m, Robert Forkel, Martin Haspelmath, and Sebastian Bank. 2023.
\newblock \href {https://glottolog.org} {Glottolog 4.8}.

\bibitem[{Jiang et~al.(2024)Jiang, Sablayrolles, Roux, Mensch, Savary, Bamford, Chaplot, de~las Casas, Bou~Hanna, Bressand, Lengyel, Bour, Lample, Lavaud, Saulnier, Lachaux, Stock, Subramanian, Yang, Antoniak, Le~Scao, Gervet, Lavril, Wang, Lacroix, and El~Sayed}]{mistral2024mixtral}
Albert~Q. Jiang, Alexandre Sablayrolles, Antoine Roux, Arthur Mensch, Blanche Savary, Chris Bamford, Devendra~Singh Chaplot, Diego de~las Casas, Emma Bou~Hanna, Florian Bressand, Gianna Lengyel, Guillaume Bour, Guillaume Lample, Lélio~Renard Lavaud, Lucile Saulnier, Marie-Anne Lachaux, Pierre Stock, Sandeep Subramanian, Sophia Yang, and 7 others. 2024.
\newblock \href {https://arxiv.org/abs/2401.04088} {Mixtral of experts}.
\newblock \emph{arXiv preprint arXiv:2401.04088}.

\bibitem[{Kocmi et~al.(2022)Kocmi, Bawden, Bojar, Dvorkovich, Federmann, Fishel, Gowda, Graham, Grundkiewicz, Haddow, Knowles, Koehn, Monz, Morishita, Nagata, Nakazawa, Novák, Popel, Popović, and Shmatova}]{kocmi2022findings}
Tom Kocmi, Rachel Bawden, Ondřej Bojar, Anton Dvorkovich, Christian Federmann, Mark Fishel, Thamme Gowda, Yvette Graham, Roman Grundkiewicz, Barry Haddow, Rebecca Knowles, Philipp Koehn, Christof Monz, Makoto Morishita, Masaaki Nagata, Toshiaki Nakazawa, Michal Novák, Martin Popel, Maja Popović, and Mariya Shmatova. 2022.
\newblock \href {https://aclanthology.org/2022.wmt-1.1/} {Findings of the 2022 conference on machine translation (wmt22)}.
\newblock In \emph{Proceedings of the Seventh Conference on Machine Translation (WMT)}, pages 1--45, Abu Dhabi, United Arab Emirates (Hybrid). Association for Computational Linguistics.

\bibitem[{Liu et~al.(2020)Liu, Gu, Goyal, Li, Edunov, Ghazvininejad, Lewis, and Zettlemoyer}]{liu2020multilingual}
Yinhan Liu, Jiatao Gu, Naman Goyal, Xian Li, Sergey Edunov, Marjan Ghazvininejad, Mike Lewis, and Luke Zettlemoyer. 2020.
\newblock \href {https://doi.org/10.1162/tacl_a_00343} {Multilingual denoising pre-training for neural machine translation}.
\newblock \emph{Transactions of the Association for Computational Linguistics}, 8:726--742.

\bibitem[{Loshchilov and Hutter(2019)}]{loshchilov2019decoupled}
Ilya Loshchilov and Frank Hutter. 2019.
\newblock \href {https://openreview.net/forum?id=Bkg6RiCqY7} {Decoupled weight decay regularization}.
\newblock In \emph{Proceedings of the 7th International Conference on Learning Representations (ICLR)}.
\newblock OpenReview preprint.

\bibitem[{Martins and Astudillo(2016)}]{martins2016from}
Andr{\'e} F.~T. Martins and Ramon Astudillo. 2016.
\newblock \href {http://proceedings.mlr.press/v48/martins16.pdf} {From softmax to sparsemax: A sparse model of attention and multi-label classification}.
\newblock In \emph{Proceedings of the 33rd International Conference on Machine Learning (ICML)}, volume~48, pages 1614--1623, New York, NY, USA. PMLR.

\bibitem[{OpenAI(2023)}]{openai2023gpt4}
OpenAI. 2023.
\newblock {GPT-4} technical report.
\newblock \url{https://openai.com/research/gpt-4}.

\bibitem[{Papineni et~al.(2002)Papineni, Roukos, Ward, and Zhu}]{papineni2002bleu}
Kishore Papineni, Salim Roukos, Todd Ward, and Wei-Jing Zhu. 2002.
\newblock Bleu: a method for automatic evaluation of machine translation.
\newblock In \emph{ACL}, pages 311--318.

\bibitem[{Potthast et~al.(2013)Potthast, Barrón-Cedeño, Stein, and Rosso}]{potthast2013overview}
Martin Potthast, Alberto Barrón-Cedeño, Benno Stein, and Paolo Rosso. 2013.
\newblock Overview of the 5th international competition on plagiarism detection.
\newblock In \emph{Working Notes for CLEF 2013 Conference}, volume 1179, pages 1--20. CEUR Workshop Proceedings.

\bibitem[{Rei et~al.(2020)Rei, Farinha, Lavie, and Specia}]{rei2020comet}
Ricardo Rei, Ana Farinha, Alon Lavie, and Lucia Specia. 2020.
\newblock Comet: A neural framework for mt evaluation.
\newblock In \emph{EMNLP}, pages 2685--2702.

\bibitem[{Rei et~al.(2023)Rei, Farinhas, Martins, and Specia}]{rei2023xcomet}
Ricardo Rei, António Farinhas, André F.~T. Martins, and Lucia Specia. 2023.
\newblock xcomet: Transparent machine translation evaluation through fine-grained error detection.
\newblock In \emph{Proceedings of the 61st Annual Meeting of the Association for Computational Linguistics (ACL)}, pages 8334--8352. Association for Computational Linguistics.

\bibitem[{Rei et~al.(2022)Rei, Treviso, Guerreiro, Zerva, Farinha, Maroti, Souza, Glushkova, Alves, Coheur, Lavie, and Martins}]{rei2022cometkiwi}
Ricardo Rei, Marcos Treviso, Nuno~M. Guerreiro, Chrysoula Zerva, Ana~C. Farinha, Christine Maroti, José G. C.~de Souza, Taisiya Glushkova, Duarte Alves, Luisa Coheur, Alon Lavie, and André F.~T. Martins. 2022.
\newblock Cometkiwi: Ist-unbabel 2022 submission for the quality estimation shared task.
\newblock In \emph{Proceedings of the Seventh Conference on Machine Translation (WMT)}. Association for Computational Linguistics.

\bibitem[{Ruder et~al.(2021)Ruder, Raganato, Staniszewski, Singh, and et~al.}]{ruder2021xtreme}
Sebastian Ruder, Alessandro Raganato, Marcin Staniszewski, Shruti Singh, and et~al. 2021.
\newblock \href {https://arxiv.org/abs/2104.07412} {{XTREME}-r: Towards more challenging and nuanced multilingual evaluation}.
\newblock \emph{arXiv preprint}, arXiv:2104.07412.

\bibitem[{Sellam et~al.(2020)Sellam, Das, and Parikh}]{sellam2020bleurt}
Thibault Sellam, Dipanjan Das, and Ankur~P. Parikh. 2020.
\newblock \href {https://doi.org/10.18653/v1/2020.acl-main.704} {{BLEURT}: Learning robust metrics for text generation}.
\newblock In \emph{Proceedings of the 58th Annual Meeting of the Association for Computational Linguistics}, pages 7881--7892. Association for Computational Linguistics.

\bibitem[{Snover et~al.(2006)Snover, Dorr, Schwartz, Micciulla, and Makhoul}]{snover2006study}
Matthew Snover, Bonnie Dorr, Richard Schwartz, Linnea Micciulla, and John Makhoul. 2006.
\newblock \href {https://aclanthology.org/2006.amta-1.36} {A study of translation edit rate with targeted human annotation}.
\newblock In \emph{Proceedings of the 7th Conference of the Association for Machine Translation in the Americas}, pages 223--231. Association for Machine Translation in the Americas.

\bibitem[{Team et~al.(2023)Team, Anil, Borgeaud, Alayrac, Yu, Soricut, Schalkwyk, Dai, Hauth, Millican et~al.}]{google2023gemini}
Gemini Team, Rohan Anil, Sebastian Borgeaud, Jean-Baptiste Alayrac, Jiahui Yu, Radu Soricut, Johan Schalkwyk, Andrew~M. Dai, Anja Hauth, Katie Millican, and 1 others. 2023.
\newblock \href {https://arxiv.org/abs/2312.11805} {Gemini: A family of highly capable multimodal models}.
\newblock \emph{arXiv preprint arXiv:2312.11805}.

\bibitem[{Team et~al.(2022)Team, Costa-juss{\`a}, Cross et~al.}]{nllbteam2022}
NLLB Team, Marta~R. Costa-juss{\`a}, James Cross, and 1 others. 2022.
\newblock No language left behind: Scaling human-centered machine translation.
\newblock \emph{arXiv preprint arXiv:2207.04672}.

\bibitem[{Thompson et~al.(2024)Thompson, Mathur, Deutsch, and Khayrallah}]{thompson2024improving}
Brian Thompson, Nitika Mathur, Daniel Deutsch, and Huda Khayrallah. 2024.
\newblock Improving statistical significance in human evaluation of automatic metrics via soft pairwise accuracy.
\newblock \emph{arXiv preprint arXiv:2409.09598}.

\bibitem[{Tuan et~al.(2021)Tuan, El-Kishky, Renduchintala, Chaudhary, Guzm{\'a}n, and Specia}]{tuan2021quality}
Yi-Lin Tuan, Ahmed El-Kishky, Adithya Renduchintala, Vishrav Chaudhary, Francisco Guzm{\'a}n, and Lucia Specia. 2021.
\newblock \href {https://aclanthology.org/2021.eacl-main.51} {Quality estimation without human-labeled data}.
\newblock In \emph{Proceedings of the 16th Conference of the European Chapter of the Association for Computational Linguistics: Main Volume}, pages 619--625, Online. Association for Computational Linguistics.

\bibitem[{Vaswani et~al.(2017)Vaswani, Shazeer, Parmar, Uszkoreit, Jones, Gomez, Kaiser, and Polosukhin}]{vaswani2017attention}
Ashish Vaswani, Noam Shazeer, Niki Parmar, Jakob Uszkoreit, Llion Jones, Aidan~N. Gomez, {\L}ukasz Kaiser, and Illia Polosukhin. 2017.
\newblock \href {https://papers.nips.cc/paper_files/paper/2017/hash/3f5ee243547dee91fbd053c1c4a845aa-Abstract.html} {Attention is all you need}.
\newblock In \emph{Advances in Neural Information Processing Systems}, pages 5998--6008. Curran Associates, Inc.

\bibitem[{Wan et~al.(2022)Wan, Liu, Yang, Zhang, Chen, Wong, and Chao}]{wan-etal-2022-unite}
Yu~Wan, Dayiheng Liu, Baosong Yang, Haibo Zhang, Boxing Chen, Derek Wong, and Lidia Chao. 2022.
\newblock \href {https://doi.org/10.18653/v1/2022.acl-long.558} {{U}ni{TE}: Unified translation evaluation}.
\newblock In \emph{Proceedings of the 60th Annual Meeting of the Association for Computational Linguistics (Volume 1: Long Papers)}, pages 8117--8127, Dublin, Ireland. Association for Computational Linguistics.

\end{thebibliography}
\end{document}